%% file: acl_latex.tex
\pdfoutput=1

\documentclass[11pt]{article}
\usepackage{acl}

\usepackage{times}
\usepackage{latexsym}

\usepackage[T1]{fontenc}

\usepackage[utf8]{inputenc}

\usepackage{microtype}

\usepackage{inconsolata}

\usepackage{todonotes}
\usepackage{blindtext}
\usepackage{enumitem}
\usepackage{amsmath}

\title{\textsc{SemScore}: Automated Evaluation of Instruction-Tuned LLMs based on Semantic Textual Similarity}

\author{Ansar Aynetdinov \\
  Humboldt-Universität zu Berlin \\
  \texttt{aynetdia@hu-berlin.de} \\\And
  Alan Akbik \\
  Humboldt-Universität zu Berlin \\
  \texttt{alan.akbik@hu-berlin.de} \\}

\begin{document}
\maketitle
\begin{abstract}
Instruction-tuned Large Language Models (LLMs) have recently showcased remarkable advancements in their ability to generate fitting responses to natural language instructions. However, many current works rely on manual evaluation to judge the quality of generated responses. Since such manual evaluation is time-consuming, it does not easily scale to the evaluation of multiple models and model variants. In this short paper, we propose a straightforward but remarkably effective evaluation metric called \textsc{SemScore}, in which we directly compare model outputs to gold target responses using semantic textual similarity (STS). We conduct a comparative evaluation of the model outputs of 12 prominent instruction-tuned LLMs using 8 widely-used evaluation metrics for text generation. We find that our proposed \textsc{SemScore} metric outperforms all other, in many cases more complex, evaluation metrics in terms of correlation to human evaluation. These findings indicate the utility of our proposed metric for the evaluation of instruction-tuned LLMs.
\end{abstract}

\section{Introduction}

Instruction-tuning~\cite{wei2022flan} has enabled large language models (LLMs) to produce fitting natural language responses to natural language instructions. Since the release of InstructGPT~\cite{instruct-gpt} and ChatGPT, an array of other language models~\cite{bloom, llama, falcon40b} and their instruction-tuned variants~\cite{alpaca, iyer2023optiml, vicuna2023} have emerged.

While human evaluation remains the gold standard for judging the quality of generated responses, it is time-consuming and does not easily scale to the evaluation of many models and model variants.
Given the current pace of development in the field, the need for effective automated evaluation approaches becomes apparent. 

However, as Table~\ref{tab:intro_example} illustrates with an example \textit{instruction} from the dataset of~\citet{wang2023selfinstruct}, common evaluation metrics for text generation may correlate poorly with human judgment. In this example, the \textit{model response} is given the highest possible rating by a human annotator but receives low BLEU and ROUGE-L scores due to low lexical overlap between the model and the \textit{target response} in the evaluation dataset.

\input{tables/example_intro}

\begin{figure*}[ht!]
    \vspace{-4mm}
    \centering
    \includegraphics[scale=0.85, trim=0.cm 0cm 0cm 1cm]{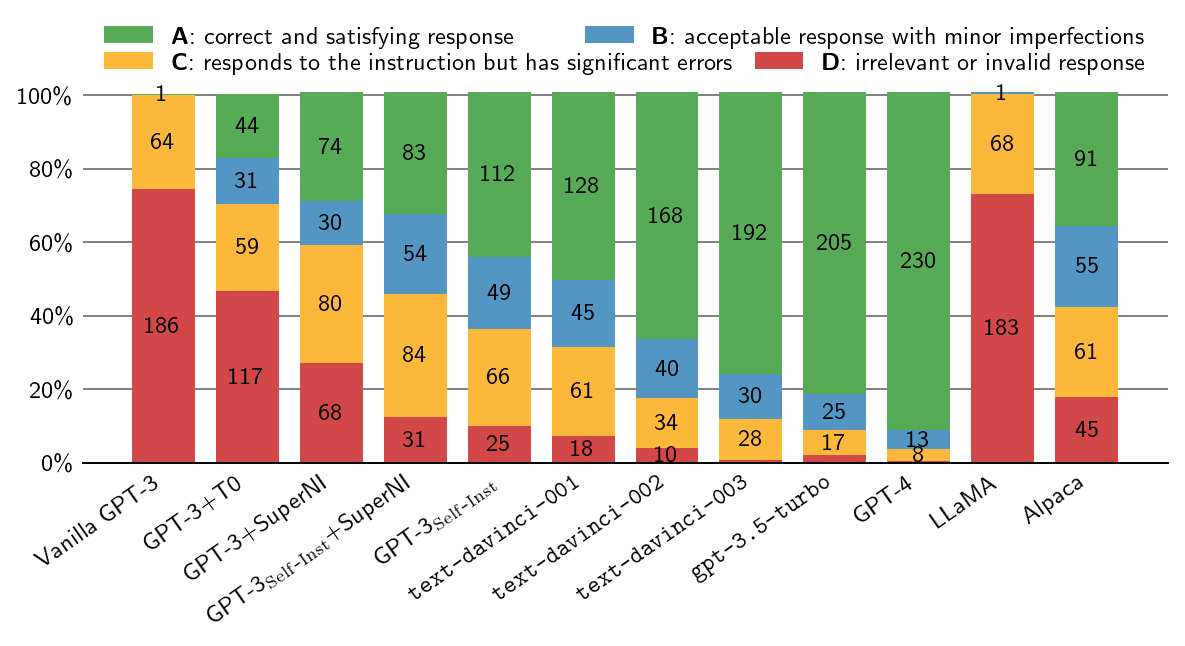}
    \vspace{-6mm}
    \caption{
    Human evaluation of prominent LLMs, based on our study and the results of \citet{wang2023selfinstruct}. From this, we derive a human-judged ranking of LLMs as basis for comparison of automated evaluation metrics.}
    \vspace{-2mm}
    \label{fig:manual_evaluation}
\end{figure*}

More generally, we observe several challenges to the automated evaluation of model responses. \textit{(1)} Traditional metrics like BLEU \cite{papineni-etal-2002-bleu} or ROUGE~\cite{lin-2004-rouge} are based on N-gram overlaps, and generally require more than one gold response, whereas instruction-tuning datasets usually contain only one target response for a given instruction~\cite{alpaca, wang-etal-2022-super, peng2023instruction}. \textit{(2)} Instruction-tuning datasets often contain coding questions, where the target answer is a code snippet, and are thus ill-suited for evaluation metrics based on N-gram overlaps or word-level embeddings. \textit{(3)} Finally, given the already large number of existing evaluation metrics for text generation~\cite{banerjee-lavie-2005-meteor, shimanaka-etal-2018-ruse, rei-etal-2020-comet, BERTScore, BARTScore}, it is unclear which best correlates with human judgement, complicating the choice of the appropriate metric. 

\noindent 
\textbf{Contributions.} To address these challenges, this short paper makes the following contributions:
\vspace{-2mm}

\begin{enumerate}[leftmargin=\parindent]
\setlength\itemsep{0em}
    \item We conduct a study in which we extend an earlier manual evaluation of model outputs of 8 instruction-tuned GPT-3 versions to include four additional models: GPT-4~\cite{gpt4}, GPT-3.5~\cite{instruct-gpt},  LLaMA~\cite{llama}, and Alpaca~\cite{alpaca}. 

    \item Based on this extended study, we produce a human-judged ranking of all 12 models.
    
    \item We evaluate 8 existing text generation metrics to determine which best correlates to this human-judged ranking.

    \item We propose an evaluation metric based on semantic textual similarity (STS) we name \textsc{SemScore}, and comparatively evaluate it against the 8 aforementioned metrics. 

\end{enumerate}

\vspace{-2mm}
We find that \textsc{SemScore} correlates best with human judgement, indicating its usefulness for automated evaluation. Furthermore, we argue that the conceptual simplicity of the method makes it well-suited to practical application.

\section{Human-Judged Ranking of LLMs}

Our first step is to compile a ranking of prominent LLMs based on human judgment. This ranking serves as basis of comparison for the automatic evaluation methods we consider in Section~\ref{sec:comparison}. 

\noindent
\textbf{Dataset.} We use the evaluation dataset of \citet{wang2023selfinstruct}. It consists of 252 instructions that, instead of focusing on traditional NLP tasks like e.g. text summarization or classification, cover a variety of tasks ranging from text completion and blog post suggestions to coding and formal logic problems, motivated by real world use cases. 
They used these instructions to manually evaluate GPT-3 \cite{gpt3} and 7 of its instruction-tuned variants. The corresponding model responses, together with target responses written by human experts, were released by the authors. We use these target responses as gold references for our manual ranking, and to calculate evaluation metrics in Section~\ref{sec:comparison}. 

\noindent
\textbf{Additional LLMs.} 
We manually evaluate four additional popular LLMs: \textit{(1)} GPT-4, \textit{(2)} GPT-3.5 model "\texttt{gpt-3.5-turbo}", \textit{(3)} LLaMA, and \textit{(4)} Alpaca-tuned LLaMA. For GPT-4 and GPT-3.5, we use the same generation parameters as \citet{wang2023selfinstruct}. For Alpaca, we reproduce the fine-tuning of LLaMA using the code by \citet{alpaca} and apply greedy decoding during inference.

\noindent
\textbf{Evaluation.} 
We follow a four-category rating system defined in~\citet{wang2023selfinstruct} that rates response on a scale from A (best rating) to D (lowest rating).
The majority of our evaluation was carried out by one human expert, while the second human expert evaluated a sample of the generated sequences, in order to validate the scores of the former one. Further details regarding the annotators can be found in Appendix \ref{appendix:annotators}.

\input{tables/ranking}

\noindent
\textbf{Results (Figure~\ref{fig:manual_evaluation}).}
The results of our manual evaluation, combined with the results of \citet{wang2023selfinstruct}, are shown in Figure~\ref{fig:manual_evaluation}.

We find GPT-4 to outperform all other models in consideration, with GPT-3.5 a close second. Unsurprisingly, both base LLMs (vanilla GPT-3 and LLaMa) score very low, as they are not instruction-tuned. Despite having only 7B trainable parameters, we find Alpaca to be comparable to the 175B parameter GPT-variant GPT-3$_{\mathrm{Self}\textnormal{-}\mathrm{Inst}}$+SuperNI.

\section{Comparison of Evaluation Metrics}
\vspace{-1mm}

We evaluate 8 popular evaluation metrics, and propose a simple additional metric we call \textsc{SemScore}, to ascertain which evaluation metric best correlates with human judgment.

\label{sec:comparison}
\subsection{Baseline Metrics}

We consider the following widely-used metrics:

\def\fs{\kern 0.33em}

\vspace{1.1mm}
\noindent
\textbullet\fs\textbf{\textbf{BLEU} {\normalfont\cite{sacrebleu}} \& \textbf{ROUGE-L} {\normalfont\cite{lin-2004-rouge}}} are traditional metrics that take into account N-gram overlaps between a candidate sequence and ideally a variety of reference sequences. We use ROUGE-L implementation provided by Google Research.

\vspace{1.1mm}
\noindent
\textbullet\fs
\textbf{\textbf{BERTScore} {\normalfont\cite{BERTScore}}} computes the similarity of two sequences as the sum of cosine similarities between their transformer-generated token embeddings. We use \texttt{deberta-xlarge-mnli} \cite{he2021deberta} as the backbone model, which currently shows the strongest correlation to human judgement, as reported by \citet{BERTScore}.

\vspace{1.1mm}
\noindent
\textbullet\fs 
\textbf{\textbf{BLEURT} {\normalfont\cite{sellam-etal-2020-bleurt}}}, a reference-free learned evaluation metric based on BERT, additionally pre-trained on Wikipedia-based synthetic data augmented with supervision signals like BLEU or BERTScore, and fine-tuned on human-rated data.

\vspace{1.1mm}
\noindent
\textbullet\fs
\textbf{\textbf{BARTScore}} \cite{BARTScore} relies on the log probability of a target sequence given a reference sequence, calculated with a pre-trained BART model \cite{lewis-etal-2020-bart}. Following \citet{BARTScore}, we use BART fine-tuned on \texttt{CNNDM} dataset \cite{hermann2015cnndm}. We additionally evaluate include \textbf{BARTScore$_{para}$}, which was fine-tuned on \texttt{ParaBank2}~\cite{hu-etal-2019-large}.

\vspace{1.1mm}
\noindent
\textbullet\fs 
\textbf{\textbf{DiscoScore}} \cite{zhao-etal-2023-discoscore}, a recently proposed BERT-based metric focusing on the discourse coherence of generated sequences. We use its best reported version DS-FOCUS (NN).

\vspace{1.1mm}
\noindent
\textbullet\fs 
\textbf{\textbf{G-Eval}} \cite{liu2023geval} is a recently proposed approach that leverages LLMs and prompting to evaluate the quality of generated texts. We use a prompt created following examples from \citet{liu2023geval} (see Appendix \ref{sec:appendix_prompt}). Since the choice of LLM is crucial to evaluation results, we evaluate G-Eval with three different LLMs: \textit{(1)} The setup designated "G-Eval-4" uses GPT-4 as backbone. \textit{(2)} The setup "G-Eval-3.5" uses \texttt{gpt-3.5-turbo}. \textit{(3)} The setup "G-Eval-3.5-instruct" uses \texttt{gpt-3.5-turbo-instruct}. We exclude those model responses that were generated by the same model used as a backbone for G-Eval in order to avoid self-evaluation.

\vspace{-1mm}
\subsection{Proposed Metric}

We additionally propose \textsc{SemScore} as a direct application of semantic textual similarity: it computes the similarity of a model response to a target response as the similarity of their respective embeddings. It consists of two steps: 
\textit{(1)} We embed model and target response separately using the current best available sentence transformer~\cite{reimers-gurevych-2019-sentence}, namely \texttt{all-mpnet-base-v2}. This model is based on MPNet-Base~\cite{MPNet}, fine-tuned with a contrastive objective on a dataset of one billion sentence pairs spanning various domains. \textit{(2)} We compute the cosine similarity of the respective embeddings as the value that constitutes the \textsc{SemScore}. 

This value lies within the interval of $[-1, 1]$. If the cosine similarity between two sequence embeddings is closer to 1, it implies two semantically similar sequences, 
while negative values imply semantically opposite sequences.
This property of cosine similarity makes \textsc{SemScore} an easily interpretable metric.
\vspace{-1mm}

\subsection{Comparing Rankings}

Table~\ref{tab:ranking} reports a ranking of all 12 models computed from human judgement and all 9 considered evaluation metrics. To compute the human ranking, we convert the human-assigned categories A-D into scores 1-4 (lower is better) and then average those scores over the entire dataset. The rankings of the 9 evaluation metrics are similarly computed by averaging them over the whole dataset. 

In order to quantify the degree of correlation between the averaged human scores and automated metric values, we calculate the Kendall rank correlation $\tau$ and the Pearson correlation coefficient $r$. Refer to Table~\ref{tab:corr_human} for the resulting correlation values.

\vspace{-1mm}

\input{tables/corr_human}

\subsection{Results and Discussion}
\label{sec:results}

As Table~\ref{tab:corr_human} shows, we find that \textsc{SemScore}, G-Eval and BERTScore show the strongest correlation to human judgement. 

\noindent 
\textbf{Limitations of LLM-based metrics.} For G-Eval, we note that its two best-scoring configurations necessitated excluding one transformer from the evaluation, meaning that only G-Eval-3.5-instruct is directly comparable to the other metrics. Still, we find that this LLM-based metric correlates strongly with human judgment, indicating the viability of using LLMs as evaluators.

\noindent 
\textbf{Evaluation using sentence embeddings.} Among the two embedding-based approaches, we note that \textsc{SemScore} slightly outperforms BERTScore, despite the smaller size of the underlying transformer \footnote{BERTScore's \texttt{deberta-xlarge-mnli} has 48 layers with a hidden layer size of 1024, while \texttt{all-mpnet-base-v2} has only 12 layers and a hidden layer size of 768.}. To gain more insight, we conduct an ablation experiment in which both metrics use the same transformer, namely BERTScore's \texttt{deberta-xlarge-mnli}. For \textsc{SemScore}, we experiment with using the CLS token and a mean mean pooling over all token embeddings to produce a sentence representation. As Table~\ref{tab:corr_human_ablation} shows, \textsc{SemScore} compares favorably even with a transformer not specifically trained for sentence embeddings.
This indicates the viability of using sentence-level representations for evaluation. 
\vspace{-1mm}

\input{tables/corr_human_ablation}

\section{Related Work}
\vspace{-1mm}

Next to the LLM-based metric G-Eval, which was considered in this paper, a number of other works proposed leveraging LLMs themselves as proxies for human evaluators~\cite{fu2023gptscore,vicuna2023,zhou2023lima}. Following this line of work, \citet{chia2023instructeval} introduced a benchmark for evaluation of instruction-tuned LLMs.

However, \citet{wang2023large} conducted a study to analyse LLM-based evaluation approaches, and found that LLMs can be manipulated through choice of prompting to influence the evaluation score, which shows that LLMs are prone to positional bias. The issue of LLM biases in general has been discussed extensively \cite{li2023survey, wan2023kelly, Kotek_2023, haller2023opiniongpt}, and this raises concerns with regards to evaluation of open-ended instruction completions that require a certain level of world knowledge. At the time of writing, there is still no clear consensus on a fully reliable setup of LLM-based evaluations~\cite{wang2023chatgpt}.

Furthermore, these approaches often rely on access to GPT-4 as "evaluator". This raises issues of reproducibility, as GPT-4 is proprietary and may be updated over time. By contrast, the evaluation metrics considered in this paper do not rely on such access or prompt engineering.

\vspace{-1mm}
\section{Conclusion}
\vspace{-1mm}

In this paper, we addressed the challenge of evaluating the quality of responses generated by instruction-tuned LLMs. 
We compared 8 widely-used evaluation metrics for text generation, and proposed a simple new metric based on textual similarity, in terms of correlation to human judgment. We find that \textsc{SemScore}  exhibits the strongest correlation to human evaluation results, even outperforming LLM-based metrics, while not requiring any special access or incurring additional costs. This indicates that \textsc{SemScore} may offer a straightforward, reproducible and cost-effective way of evaluating the quality of LLM responses.

\section*{Limitations}

One limitation of \textsc{SemScore} is its dependence on an underlying transformer model to compute semantic textual similarity between model and target outputs. In this paper, we used the strongest currently-available sentence transformer model \texttt{all-mpnet-base-v2}. However, future research might yield improved STS models that may produce different similarity scores. Our ablation presented in Section~\ref{sec:results} indeed shows that the choice of transformer influences results. To mitigate this risk, we make a clear recommendation in this paper to use \texttt{all-mpnet-base-v2}, and advise all future works that report \textsc{SemScore} to either use this model or name the underlying transformer model they use. 

In addition, a more general limitation is that \textsc{SemScore} (like all other metrics considered in this paper) requires at least one gold-standard target output against which to compare a generated response. This target output should be human created or at least human-vetted. 

Lastly, we acknowledge the small size of the evaluation dataset used in our experiments and its lack of focus on traditional NLP tasks, however argue that this dataset is realistic for evaluation of instruction-tuned LLMs, as end users might use them for a broad range of text-oriented tasks, going beyond the traditional NLP ones.

\bibliography{acl_latex}
\bibliographystyle{acl_natbib}

\appendix

\section{Appendix}
\label{sec:appendix}

\subsection{Per-task correlations}
\label{sec:appendix_pertask_corr}

In order to provide additional understanding of how each metric deals with various types of tasks in our dataset, we provide a further per-task breakdown in Table \ref{tab:group_corr}. We provide correlations only for those instruction groups that have at least 6 instances over the four model variants we evaluated manually (\texttt{gpt-3.5-turbo}, GPT-4, LLaMA and Alpaca). Evaluation scores for other models were reported by \cite{wang2023selfinstruct} on an aggregated level, and thus could not be considered in this breakdown. For G-Eval-4 we excluded GPT-4 outputs. We find variations in scores for different metrics across different task groups, but find that overall, SemScore compares favorably.

\subsection{Examples}
\label{sec:appendix_examples}

Table~\ref{tab:examples} presents examples of instructions and their target responses, as well as the model response together with its human-assigned rating and evaluation scores by the top 3 performing metrics in our evaluation. We chose short examples for space reasons, though many instructions and responses in the dataset are rather lengthy. 

We discuss the 5 examples listed in Table~\ref{tab:examples}: 

\noindent 
\textbf{Example 1} illustrates a creative task that shows the strengh of embedding-based evaluations. While \textsc{SemScore} a high score to the model response, matching human rating, BERTScore is slightly more moderate. G-Eval-4 gives an appropriate score as well. ROUGE-L, however, is not able to detect matching semantic structures in the model and target responses and thus incorrectly assigns a very low score.

\noindent 
\textbf{Example 2} is a syntax-heavy task in which ROUGE-L is able to compete with \textsc{SemScore}, while BERTScore is a little too high in this case. G-Eval-4 in this case fails to score the response appropriately.

\noindent 
\textbf{Example 3} shows a coding-related instruction in which \textsc{SemScore} scores surprisingly badly given that it should be able to detect similar code structures based on the data that it was fine-tuned on. However, it is possible that the natural language comment in the target overly affects the generated embedding, as transformers like \texttt{all-mpnet-base-v2} were shown to be biased towards noun participants (like fruit names in this case) \cite{nikolaev-pado-2023-representation}. G-Eval-4, in this instance, fares best compared to other metrics.

\noindent 
\textbf{Example 4 and 5} illustrate responses that should not pose a big challenge for any of the metrics. However, BERTScore assigns an unfitting score to example 4. In example 5, \textsc{SemScore} does not penalize the extra genre enough, together with with G-Eval-4, and ROUGE-L does so perhaps too much.

\input{tables/task_corr}
\input{tables/examples}

\subsection{G-Eval Prompt}

The prompt for G-Eval is based on the example prompts provided by \citet{liu2023geval} and the score descriptions provided by \citet{wang2023selfinstruct}:

\begin{quote}
\label{sec:appendix_prompt}
{\itshape
You will be given an instruction-output pair. Your task is to rate the responses on one metric.

Please make sure you read and understand these instructions carefully. Please keep this document open while reviewing, and refer to it as needed.

Evaluation Criteria:\\
Overall Quality (1-4) - how well does the output complete the instruction?\\
- A score of 1 means that the response is valid and satisfying. It follows the instruction, properly completes it and does not contain any repetitions or irrelevant parts.\\
- A score of 2 means that the response is acceptable but has minor errors or imperfections. It may contain factual inconistensies or grammatical errors.\\
- A score of 3 means that the response is relevant and responds to the instruction, but it has significant errors in the content. For example, the output may be valid in the beginning, but contains repetitions or followed by irrelevant things afterwards.\\
- A score of 4 means that the response is irrelevant or completely invalid, i.e. consists of repeating sequences or does not correspond to the instruction in any way.

Evaluation Steps:\\
1. Read the instruction and the corresponding output carefully.\\
2. Rate the output on a scale of 1-4 for Quality, according to the criteria above.\\

\#\#\# Instruction:\\
\{\{Instruction\}\}\\
\#\#\# Corresponding Output:\\
\{\{Output\}\}\\

Evaluation Form (scores ONLY): \\
- Quality:
}
\end{quote}

\subsection{Human annotators}
\label{appendix:annotators}

The results of human evaluation suggest a strong agreement between experts with a Kappa score of 0.63. The experts followed the setup described by \citet{wang2023selfinstruct} in their evaluation. As for the experts' background, they are computer science PhD students, aged under 30, Europeans, white males. They were compensated for their work in accordance with their employment contracts.

\end{document}

%% file: tables/example_intro.tex
\setlength\tabcolsep{1 pt} 
\newcommand{\sty}{\tt \small}

\setlist{leftmargin=2mm}

\setlength{\tabcolsep}{5pt}
\begin{table}[t]
    \vspace{-5mm}
    \centering
    \resizebox{0.47\textwidth}{!}{
    \begin{tabular}{p{0.16\textwidth} p{0.3\textwidth}}
    \hline
    \hline
    \sty
     Instruction: & \sty Give some examples of what people usually say  when someone arrives safely \\
     \sty
     Target Response: & \sty Glad you made it safe and sound. \\
     \sty
     Model Response: & \sty Thank goodness you arrived without any issues. \\
     \sty ROUGE-L: & \sty \textcolor{red}{0.143} \\
     \sty BLEU: & \sty \textcolor{red}{6.57} \\
     \sty Human Rating: & \sty \textbf{\textcolor{ForestGreen}{A} (best)} \\
    \hline
    \hline
    \end{tabular}
    }
    \caption{An example in which a generated \textit{model response} to an \textit{instruction} is rated as very high quality (\textcolor{ForestGreen}{\textbf{A}}-rating) by human evaluation, but scores a very low BLEU score due to low N-gram overlap to the gold reference \textit{target response}.}
    \vspace{-5mm}
\label{tab:intro_example}
\end{table}

%% file: tables/ranking.tex
\begin{table*}[ht]
\vspace{-1mm}
    \centering
    \resizebox{0.99\textwidth}{!}{
    \begin{tabular}{lcccccccccc}
    \hline
    \hline
    Model & Human & \textsc{SemScore} & \textit{G-Eval-4}$^*$ & BERTScore & ROUGE-L & BARTScore & BARTScore$_{para}$ & BLEU & BLEURT & DiscoScore \\ 
    \hline
    GPT-4 & 1 & 1 &  & 2 & 3 & 2 & 2 & 6 & 2 & 5  \\
    \texttt{gpt-3.5-turbo} & 2 & 4 & 1 & 4 & 5 & 3 & 5 & 5 & 4 & 3 \\
    \texttt{text-davinci-003} & 3 & 2 & 2 & 1 & 1 & 1 & 1 & 3 & 3 & 8 \\
    \texttt{text-davinci-002} & 4 & 3 & 3 & 3 & 2 & 4 & 3 & 2 & 5 & 4 \\
    \texttt{text-davinci-001} & 5 & 5 & 4 & 5 & 4 & 6 & 7 & 1 & 6 & 2 \\
    GPT-3$_{\mathrm{Self}\textnormal{-}\mathrm{Inst}}$ & 6 & 7 & 5 & 7 & 6 & 7 & 4 & 7 & 8 & 9 \\
    Alpaca & 7 & 6 & 11 & 6 & 8 & 5 & 8 & 4 & 9 & 1\\
    GPT-3$_{\mathrm{Self}\textnormal{-}\mathrm{Inst}}$+SuperNI & 8 & 8 & 6 & 8 & 7 & 8 & 6 & 8 & 11 & 7 \\
    GPT-3+SuperNI & 9 & 9 & 7 & 9 & 9 & 9 & 9 & 9 & 10 & 11\\
    GPT-3+T0 & 10 & 10 & 8 & 10 & 10 & 12 & 12 & 10 & 12 & 6 \\
    Vanilla GPT-3 & 11 & 12 & 9 & 12 & 12 & 11 & 11 & 12 & 1 & 12 \\
    LLaMA & 12 & 11 & 10 & 11 & 11 & 10 & 10 & 11 & 7 & 10 \\
    \hline
    \hline
    \end{tabular}
    }
\vspace{-1mm}
    \caption{Ranking of each model from best (rank 1) to worst (rank 12) according to each metric we consider. Since we use GPT-4 as backbone for G-Eval, we exclude GPT-4 outputs from its evaluation.}
\label{tab:ranking}
\vspace{-4mm}
\end{table*}

%% file: tables/corr_human.tex
\begin{table}[t]
    \centering
    \begin{tabular}{lcc}
    \hline
    \hline
    Metric & $\tau$ & $r$ \\ 
    \hline
    \textsc{SemScore} & \textbf{0.879} & \textbf{0.970} \\
    \textit{G-Eval-4}$^*$ & 0.855 & 0.863 \\
    \textit{G-Eval-3.5}$^*$ & 0.855 & 0.831 \\
    BERTScore & 0.848 & 0.944 \\
    G-Eval-3.5-instruct & 0.840 & 0.911 \\
    ROUGE-L & 0.788 & 0.933 \\
    BARTScore & 0.788 & 0.621 \\
    BARTScore$_{para}$ & 0.697 & 0.884 \\
    BLEU & 0.667 & 0.865 \\
    BLEURT & 0.485 & 0.485\\
    DiscoScore & 0.364 & 0.583\\
    \hline
    \hline
    \end{tabular}
    \caption{Kendall $\tau$   \& Pearson $r$ correlation (absolute values) of averaged automated evaluation metrics to averaged human scores. We excluded the evaluations of corresponding GPT models when calculating correlation values for G-Eval scores noted with $*$.}
    \vspace{-4mm}
\label{tab:corr_human}
\end{table}

%% file: tables/corr_human_ablation.tex
\begin{table}[t]
    \centering
    \begin{tabular}{lcc}
    \hline
    \hline
    Metric & $\tau$ & $r$ \\ 
    \hline
    \textsc{SemScore} & \textbf{0.879} & \textbf{0.970} \\
    \textsc{SemScore}$_{\mathrm{DeBERTa}\textnormal{-}\mathrm{Mean}}$ & 0.870 & 0.929 \\
    BERTScore & 0.848 & 0.944 \\
    \textsc{SemScore}$_{\mathrm{DeBERTa}\textnormal{-}\mathrm{CLS}}$ & 0.756 & 0.892 \\
    \hline
    \hline
    \end{tabular}
    \caption{Correlation of \textsc{SemScore} to human evaluation, when calculated on the CLS token and mean pooling of all tokens in model outputs using \texttt{deberta-xlarge-mnli} embeddings.}
    \vspace{-3mm}
\label{tab:corr_human_ablation}
\end{table}

%% file: tables/task_corr.tex
\begin{table*}[ht]
    \centering
    \resizebox{0.99\textwidth}{!}{
    \begin{tabular}{lrrrrrrrrrrrrrrr}
    \hline
    \hline
    Metric & Grammarly & merriam-webster.com & Gmail & Netflix & Amazon & IMDB & Tasty & Leetcode & Spotify & Overleaf & tripadvisor.com & Messenger & Wikipedia & StackOverflow & Twitter \\
    \hline
    SemScore & 0.548 & 1 & 1 & 0.667 & 0.913 & 1 & 0.913 & 0.913 & 1 & 0.333 & 1 & 0.667 & 0.913 & 0.913 & 1 \\
    G-Eval-4 & 0 & 0.333 & 0.333 & 0 & 0 & 0.333 & 0.333 & 0.816 & 1 & 0 & 0.333 & 0.816 & 0 & 0.333 & 0 \\
    BERTScore & 0.548 & 1 & 1 & 0.667 & 0.913 & 1 & 0.913 & 0.913 & 1 & 1 & 1 & 0.667 & 0.913 & 0.913 & 0.667 \\
    ROUGE-L & 0.548 & 1 & 1 & 0.667 & 0.913 & 0.667 & 0.548 & 0.183 & 0.667 & 1 & 0.667 & 0.667 & 0.913 & 0.913 & 0.667 \\
    BARTScore & 0.548 & 0.333 & 0.667 & 0 & 0.913 & 1 & 0.548 & 0.548 & 0.333 & 0.333 & 0.333 & 0.333 & 0.913 & 0.913 & 1 \\
    BARTScore$_{para}$ & 0.548 & 1 & 1 & 0.667 & 0.548 & 1 & 0.913 & 0.913 & 0.333 & 1 & 0 & 1 & 0.913 & 0.913 & 1 \\
    BLEU & 0.548 & 0.333 & 0.333 & 0.333 & 0.548 & 0.667 & 0.183 & 0.913 & 0 & 0.667 & 0.667 & 0.667 & 0.548 & 0.913 & 0.333 \\
    BLEURT & 0.183 & 0 & 1 & 0 & 0.548 & 1 & 0.548 & 0.548 & 0.667 & 0 & 0 & 0.333 & 0.548 & 0.548 & 0.667 \\
    DiscoScore & 0.913 & 0.913 & 0 & 0 & 0.548 & 0.667 & 0.548 & 0.183 & 0.333 & 0 & 0.333 & 0.667 & 0.548 & 0.913 & 0.333 \\
    \hline
    \hline
    \end{tabular}
    }
    \caption{Per-task breakdown of correlations between evaluated metrics and human scores.}
\label{tab:group_corr}
\end{table*}

%% file: tables/examples.tex
\setlength\tabcolsep{1 pt} 
\renewcommand{\sty}{\tt \small}
\renewcommand\labelitemi{-} 

\setlist{leftmargin=2mm}

\newcommand{\ins}{\textbf{Instruction:}}
\newcommand{\inp}{\textbf{\newline{Input:}}}

\setlength{\tabcolsep}{7pt}
\begin{table*}[t!]
\centering
\small
\resizebox{0.99\textwidth}{!}{
\begin{tabular}{c p{0.30\textwidth} p{0.30\textwidth} p{0.4\textwidth} p{0.2\textwidth}}
\hline
\hline
\# & \multicolumn{1}{c}{\textbf{Instruction + Instance Input}}   & \multicolumn{1}{c}{\textbf{Target Response}} & \multicolumn{1}{c}{\textbf{Model Response}} & \multicolumn{1}{c}{\textbf{Metric: Score}}\\
\hline
\sty
1 & \sty \ins{} Write a social media post about the call for collaboration on a crowdsourcing project in a persuasive way. &  
\sty
Looking for collaborators on a new crowdsourcing project! This is a great opportunity to get involved in something innovative and impactful. If you're interested, please let me know. & 
\sty
Calling all innovators! We need your help to make a difference. Join us in our crowdsourcing project and help us create something amazing. Together, we can make a real impact and create something that will benefit us all. \#Crowdsourcing \#Collaboration \#Innovation &
\vspace{-5mm}
\begin{itemize}
\itemsep0em
  \item[] \textsc{SemScore}: \textcolor{ForestGreen}{0.722}
  \item[] G-Eval-4: \textcolor{ForestGreen}{1}
  \item[] BERTScore: \textcolor{orange}{0.654}
  \item[] ROUGE-L: \textcolor{red}{0.143}
  \item[] Human Score: A
\end{itemize} \\

\hline
\sty
2 & \sty \ins{} Identify all words that match the pattern given.
\inp{} H\_AR\_ &
\sty
haart, heard, hears, heart, hoard, hoary & 
\sty
HARD, HARSH, HEART
& 
\vspace{-5mm}
\begin{itemize}
\itemsep0em
  \item[] \textsc{SemScore}: \textcolor{ForestGreen}{0.423}
  \item[] G-Eval-4: \textcolor{red}{1}
  \item[] BERTScore: \textcolor{orange}{0.604}
  \item[] ROUGE-L: \textcolor{ForestGreen}{0.222}
  \item[] Human Score: C
\end{itemize} 
\vspace{-6mm}
\\

\hline
\sty
3 & \sty \ins{} You are given an algorithm or a search query about coding. Look for a code snippet related to the search query. 
\inp{} Python loop &  
\sty
\#Print each fruit in a fruit list: \newline
fruits = ["apple", "banana", "cherry"] \newline
for x in fruits: \newline
print(x) & 
\sty
for i in range(10): \newline
\hangindent=2em print(i) &
\vspace{-5mm}
\begin{itemize}
\itemsep0em
  \item[] \textsc{SemScore}: \textcolor{orange}{0.49}
  \item[] G-Eval-4: \textcolor{ForestGreen}{1}
  \item[] BERTScore: \textcolor{orange}{0.665}
  \item[] ROUGE-L: \textcolor{red}{0.25}
  \item[] Human Score: A
\end{itemize} \\

\hline
\sty
4 & \sty \ins{} You are given an algorithm or a search query about coding. Look for a code snippet related to the search query. Decide which part of speech the underlined word belongs to.
\inp{} school will \_keep through the winter &  
\sty
verb &
\sty
school will keep through the winter &
\vspace{-5mm}
\begin{itemize}
\itemsep0em
  \item[] \textsc{SemScore}: \textcolor{ForestGreen}{0.163}
  \item[] G-Eval-4: \textcolor{ForestGreen}{4}
  \item[] BERTScore: \textcolor{orange}{0.380}
  \item[] ROUGE-L: \textcolor{ForestGreen}{0}
  \item[] Human Score: D
\end{itemize} \\

\hline
\sty
5 & \sty \ins{} In what genres does the given movie or series fall?
\inp{} Manifest &  
\sty
Mystery, Sci-Fi, Drama &
\sty
Drama, Mystery, Sci-Fi, Thriller &
\vspace{-5mm}
\begin{itemize}
\itemsep0em
  \item[] \textsc{SemScore}: \textcolor{orange}{0.913}
  \item[] G-Eval-4: \textcolor{orange}{1}
  \item[] BERTScore: \textcolor{ForestGreen}{0.855}
  \item[] ROUGE-L: \textcolor{orange}{0.667}
  \item[] Human Score: B
\end{itemize} \\

\hline
\hline
\end{tabular}
}
\caption{
Examples of instruction-based tasks, on which all of the models in our study were evaluated. We list individual scores of top 3 performing metrics in terms of correlation with human ratings. Scores highlighted with \textcolor{ForestGreen}{green} align with the human score well, \textcolor{orange}{orange} indicates acceptable alignment, and \textcolor{red}{red} indicates a bad alignment.
}
\label{tab:examples}
\end{table*}

%% file: acl_latex.bbl
\begin{thebibliography}{40}
\expandafter\ifx\csname natexlab\endcsname\relax\def\natexlab#1{#1}\fi

\bibitem[{Almazrouei et~al.(2023)Almazrouei, Alobeidli, Alshamsi, Cappelli, Cojocaru, Debbah, Goffinet, Heslow, Launay, Malartic, Noune, Pannier, and Penedo}]{falcon40b}
Ebtesam Almazrouei, Hamza Alobeidli, Abdulaziz Alshamsi, Alessandro Cappelli, Ruxandra Cojocaru, Merouane Debbah, Etienne Goffinet, Daniel Heslow, Julien Launay, Quentin Malartic, Badreddine Noune, Baptiste Pannier, and Guilherme Penedo. 2023.
\newblock {Falcon-40B}: an open large language model with state-of-the-art performance.

\bibitem[{Banerjee and Lavie(2005)}]{banerjee-lavie-2005-meteor}
Satanjeev Banerjee and Alon Lavie. 2005.
\newblock \href {https://aclanthology.org/W05-0909} {{METEOR}: An automatic metric for {MT} evaluation with improved correlation with human judgments}.
\newblock In \emph{Proceedings of the {ACL} Workshop on Intrinsic and Extrinsic Evaluation Measures for Machine Translation and/or Summarization}, pages 65--72, Ann Arbor, Michigan. Association for Computational Linguistics.

\bibitem[{Brown et~al.(2020)Brown, Mann, Ryder, Subbiah, Kaplan, Dhariwal, Neelakantan, Shyam, Sastry, Askell, Agarwal, Herbert-Voss, Krueger, Henighan, Child, Ramesh, Ziegler, Wu, Winter, Hesse, Chen, Sigler, Litwin, Gray, Chess, Clark, Berner, McCandlish, Radford, Sutskever, and Amodei}]{gpt3}
Tom Brown, Benjamin Mann, Nick Ryder, Melanie Subbiah, Jared~D Kaplan, Prafulla Dhariwal, Arvind Neelakantan, Pranav Shyam, Girish Sastry, Amanda Askell, Sandhini Agarwal, Ariel Herbert-Voss, Gretchen Krueger, Tom Henighan, Rewon Child, Aditya Ramesh, Daniel Ziegler, Jeffrey Wu, Clemens Winter, Chris Hesse, Mark Chen, Eric Sigler, Mateusz Litwin, Scott Gray, Benjamin Chess, Jack Clark, Christopher Berner, Sam McCandlish, Alec Radford, Ilya Sutskever, and Dario Amodei. 2020.
\newblock \href {https://proceedings.neurips.cc/paper_files/paper/2020/file/1457c0d6bfcb4967418bfb8ac142f64a-Paper.pdf} {Language models are few-shot learners}.
\newblock In \emph{Advances in Neural Information Processing Systems}, volume~33, pages 1877--1901. Curran Associates, Inc.

\bibitem[{Chia et~al.(2023)Chia, Hong, Bing, and Poria}]{chia2023instructeval}
Yew~Ken Chia, Pengfei Hong, Lidong Bing, and Soujanya Poria. 2023.
\newblock \href {http://arxiv.org/abs/2306.04757} {Instructeval: Towards holistic evaluation of instruction-tuned large language models}.

\bibitem[{Chiang et~al.(2023)Chiang, Li, Lin, Sheng, Wu, Zhang, Zheng, Zhuang, Zhuang, Gonzalez, Stoica, and Xing}]{vicuna2023}
Wei-Lin Chiang, Zhuohan Li, Zi~Lin, Ying Sheng, Zhanghao Wu, Hao Zhang, Lianmin Zheng, Siyuan Zhuang, Yonghao Zhuang, Joseph~E. Gonzalez, Ion Stoica, and Eric~P. Xing. 2023.
\newblock \href {https://lmsys.org/blog/2023-03-30-vicuna/} {Vicuna: An open-source chatbot impressing gpt-4 with 90\%* chatgpt quality}.

\bibitem[{Fu et~al.(2023)Fu, Ng, Jiang, and Liu}]{fu2023gptscore}
Jinlan Fu, See-Kiong Ng, Zhengbao Jiang, and Pengfei Liu. 2023.
\newblock \href {http://arxiv.org/abs/2302.04166} {Gptscore: Evaluate as you desire}.

\bibitem[{Haller et~al.(2023)Haller, Aynetdinov, and Akbik}]{haller2023opiniongpt}
Patrick Haller, Ansar Aynetdinov, and Alan Akbik. 2023.
\newblock \href {http://arxiv.org/abs/2309.03876} {Opiniongpt: Modelling explicit biases in instruction-tuned llms}.

\bibitem[{He et~al.(2021)He, Liu, Gao, and Chen}]{he2021deberta}
Pengcheng He, Xiaodong Liu, Jianfeng Gao, and Weizhu Chen. 2021.
\newblock \href {https://openreview.net/forum?id=XPZIaotutsD} {Deberta: Decoding-enhanced bert with disentangled attention}.
\newblock In \emph{International Conference on Learning Representations}.

\bibitem[{Hermann et~al.(2015)Hermann, Kocisky, Grefenstette, Espeholt, Kay, Suleyman, and Blunsom}]{hermann2015cnndm}
Karl~Moritz Hermann, Tomas Kocisky, Edward Grefenstette, Lasse Espeholt, Will Kay, Mustafa Suleyman, and Phil Blunsom. 2015.
\newblock \href {https://proceedings.neurips.cc/paper_files/paper/2015/file/afdec7005cc9f14302cd0474fd0f3c96-Paper.pdf} {Teaching machines to read and comprehend}.
\newblock In \emph{Advances in Neural Information Processing Systems}, volume~28. Curran Associates, Inc.

\bibitem[{Hu et~al.(2019)Hu, Singh, Holzenberger, Post, and Van~Durme}]{hu-etal-2019-large}
J.~Edward Hu, Abhinav Singh, Nils Holzenberger, Matt Post, and Benjamin Van~Durme. 2019.
\newblock \href {https://doi.org/10.18653/v1/K19-1005} {Large-scale, diverse, paraphrastic bitexts via sampling and clustering}.
\newblock In \emph{Proceedings of the 23rd Conference on Computational Natural Language Learning (CoNLL)}, pages 44--54, Hong Kong, China. Association for Computational Linguistics.

\bibitem[{Iyer et~al.(2023)Iyer, Lin, Pasunuru, Mihaylov, Simig, Yu, Shuster, Wang, Liu, Koura, Li, O'Horo, Pereyra, Wang, Dewan, Celikyilmaz, Zettlemoyer, and Stoyanov}]{iyer2023optiml}
Srinivasan Iyer, Xi~Victoria Lin, Ramakanth Pasunuru, Todor Mihaylov, Daniel Simig, Ping Yu, Kurt Shuster, Tianlu Wang, Qing Liu, Punit~Singh Koura, Xian Li, Brian O'Horo, Gabriel Pereyra, Jeff Wang, Christopher Dewan, Asli Celikyilmaz, Luke Zettlemoyer, and Ves Stoyanov. 2023.
\newblock \href {http://arxiv.org/abs/2212.12017} {Opt-iml: Scaling language model instruction meta learning through the lens of generalization}.

\bibitem[{Kotek et~al.(2023)Kotek, Dockum, and Sun}]{Kotek_2023}
Hadas Kotek, Rikker Dockum, and David Sun. 2023.
\newblock \href {https://doi.org/10.1145/3582269.3615599} {Gender bias and stereotypes in large language models}.
\newblock In \emph{Proceedings of The ACM Collective Intelligence Conference}, CI ’23. ACM.

\bibitem[{Lewis et~al.(2020)Lewis, Liu, Goyal, Ghazvininejad, Mohamed, Levy, Stoyanov, and Zettlemoyer}]{lewis-etal-2020-bart}
Mike Lewis, Yinhan Liu, Naman Goyal, Marjan Ghazvininejad, Abdelrahman Mohamed, Omer Levy, Veselin Stoyanov, and Luke Zettlemoyer. 2020.
\newblock \href {https://doi.org/10.18653/v1/2020.acl-main.703} {{BART}: Denoising sequence-to-sequence pre-training for natural language generation, translation, and comprehension}.
\newblock In \emph{Proceedings of the 58th Annual Meeting of the Association for Computational Linguistics}, pages 7871--7880, Online. Association for Computational Linguistics.

\bibitem[{Li et~al.(2023)Li, Du, Song, Wang, and Wang}]{li2023survey}
Yingji Li, Mengnan Du, Rui Song, Xin Wang, and Ying Wang. 2023.
\newblock \href {http://arxiv.org/abs/2308.10149} {A survey on fairness in large language models}.

\bibitem[{Lin(2004)}]{lin-2004-rouge}
Chin-Yew Lin. 2004.
\newblock \href {https://aclanthology.org/W04-1013} {{ROUGE}: A package for automatic evaluation of summaries}.
\newblock In \emph{Text Summarization Branches Out}, pages 74--81, Barcelona, Spain. Association for Computational Linguistics.

\bibitem[{Liu et~al.(2023)Liu, Iter, Xu, Wang, Xu, and Zhu}]{liu2023geval}
Yang Liu, Dan Iter, Yichong Xu, Shuohang Wang, Ruochen Xu, and Chenguang Zhu. 2023.
\newblock \href {http://arxiv.org/abs/2303.16634} {G-eval: Nlg evaluation using gpt-4 with better human alignment}.

\bibitem[{Nikolaev and Pad{\'o}(2023)}]{nikolaev-pado-2023-representation}
Dmitry Nikolaev and Sebastian Pad{\'o}. 2023.
\newblock \href {https://aclanthology.org/2023.eacl-main.268} {Representation biases in sentence transformers}.
\newblock In \emph{Proceedings of the 17th Conference of the European Chapter of the Association for Computational Linguistics}, pages 3701--3716, Dubrovnik, Croatia. Association for Computational Linguistics.

\bibitem[{OpenAI(2023)}]{gpt4}
OpenAI. 2023.
\newblock \href {http://arxiv.org/abs/2303.08774} {Gpt-4 technical report}.

\bibitem[{Ouyang et~al.(2022)Ouyang, Wu, Jiang, Almeida, Wainwright, Mishkin, Zhang, Agarwal, Slama, Ray, Schulman, Hilton, Kelton, Miller, Simens, Askell, Welinder, Christiano, Leike, and Lowe}]{instruct-gpt}
Long Ouyang, Jeff Wu, Xu~Jiang, Diogo Almeida, Carroll~L. Wainwright, Pamela Mishkin, Chong Zhang, Sandhini Agarwal, Katarina Slama, Alex Ray, John Schulman, Jacob Hilton, Fraser Kelton, Luke Miller, Maddie Simens, Amanda Askell, Peter Welinder, Paul Christiano, Jan Leike, and Ryan Lowe. 2022.
\newblock \href {http://arxiv.org/abs/2203.02155} {Training language models to follow instructions with human feedback}.

\bibitem[{Papineni et~al.(2002)Papineni, Roukos, Ward, and Zhu}]{papineni-etal-2002-bleu}
Kishore Papineni, Salim Roukos, Todd Ward, and Wei-Jing Zhu. 2002.
\newblock \href {https://doi.org/10.3115/1073083.1073135} {{B}leu: a method for automatic evaluation of machine translation}.
\newblock In \emph{Proceedings of the 40th Annual Meeting of the Association for Computational Linguistics}, pages 311--318, Philadelphia, Pennsylvania, USA. Association for Computational Linguistics.

\bibitem[{Peng et~al.(2023)Peng, Li, He, Galley, and Gao}]{peng2023instruction}
Baolin Peng, Chunyuan Li, Pengcheng He, Michel Galley, and Jianfeng Gao. 2023.
\newblock Instruction tuning with gpt-4.
\newblock \emph{arXiv preprint arXiv:2304.03277}.

\bibitem[{Post(2018)}]{sacrebleu}
Matt Post. 2018.
\newblock \href {https://www.aclweb.org/anthology/W18-6319} {A call for clarity in reporting {BLEU} scores}.
\newblock In \emph{Proceedings of the Third Conference on Machine Translation: Research Papers}, pages 186--191, Belgium, Brussels. Association for Computational Linguistics.

\bibitem[{Rei et~al.(2020)Rei, Stewart, Farinha, and Lavie}]{rei-etal-2020-comet}
Ricardo Rei, Craig Stewart, Ana~C Farinha, and Alon Lavie. 2020.
\newblock \href {https://doi.org/10.18653/v1/2020.emnlp-main.213} {{COMET}: A neural framework for {MT} evaluation}.
\newblock In \emph{Proceedings of the 2020 Conference on Empirical Methods in Natural Language Processing (EMNLP)}, pages 2685--2702, Online. Association for Computational Linguistics.

\bibitem[{Reimers and Gurevych(2019)}]{reimers-gurevych-2019-sentence}
Nils Reimers and Iryna Gurevych. 2019.
\newblock \href {https://doi.org/10.18653/v1/D19-1410} {Sentence-{BERT}: Sentence embeddings using {S}iamese {BERT}-networks}.
\newblock In \emph{Proceedings of the 2019 Conference on Empirical Methods in Natural Language Processing and the 9th International Joint Conference on Natural Language Processing (EMNLP-IJCNLP)}, pages 3982--3992, Hong Kong, China. Association for Computational Linguistics.

\bibitem[{Scao et~al.(2023)Scao, Fan, Akiki, Pavlick, Ilić, Hesslow, Castagné, Luccioni, Yvon, Gallé, Tow, Rush, Biderman, Webson, Ammanamanchi, Wang, Sagot et~al.}]{bloom}
BigScience Workshop: Teven~Le Scao, Angela Fan, Christopher Akiki, Ellie Pavlick, Suzana Ilić, Daniel Hesslow, Roman Castagné, Alexandra~Sasha Luccioni, François Yvon, Matthias Gallé, Jonathan Tow, Alexander~M. Rush, Stella Biderman, Albert Webson, Pawan~Sasanka Ammanamanchi, Thomas Wang, Benoît Sagot, et~al. 2023.
\newblock \href {http://arxiv.org/abs/2211.05100} {Bloom: A 176b-parameter open-access multilingual language model}.

\bibitem[{Sellam et~al.(2020)Sellam, Das, and Parikh}]{sellam-etal-2020-bleurt}
Thibault Sellam, Dipanjan Das, and Ankur Parikh. 2020.
\newblock \href {https://doi.org/10.18653/v1/2020.acl-main.704} {{BLEURT}: Learning robust metrics for text generation}.
\newblock In \emph{Proceedings of the 58th Annual Meeting of the Association for Computational Linguistics}, pages 7881--7892, Online. Association for Computational Linguistics.

\bibitem[{Shimanaka et~al.(2018)Shimanaka, Kajiwara, and Komachi}]{shimanaka-etal-2018-ruse}
Hiroki Shimanaka, Tomoyuki Kajiwara, and Mamoru Komachi. 2018.
\newblock \href {https://doi.org/10.18653/v1/W18-6456} {{RUSE}: Regressor using sentence embeddings for automatic machine translation evaluation}.
\newblock In \emph{Proceedings of the Third Conference on Machine Translation: Shared Task Papers}, pages 751--758, Belgium, Brussels. Association for Computational Linguistics.

\bibitem[{Song et~al.(2020)Song, Tan, Qin, Lu, and Liu}]{MPNet}
Kaitao Song, Xu~Tan, Tao Qin, Jianfeng Lu, and Tie-Yan Liu. 2020.
\newblock \href {https://proceedings.neurips.cc/paper_files/paper/2020/file/c3a690be93aa602ee2dc0ccab5b7b67e-Paper.pdf} {Mpnet: Masked and permuted pre-training for language understanding}.
\newblock In \emph{Advances in Neural Information Processing Systems}, volume~33, pages 16857--16867. Curran Associates, Inc.

\bibitem[{Taori et~al.(2023)Taori, Gulrajani, Zhang, Dubois, Li, Guestrin, Liang, and Hashimoto}]{alpaca}
Rohan Taori, Ishaan Gulrajani, Tianyi Zhang, Yann Dubois, Xuechen Li, Carlos Guestrin, Percy Liang, and Tatsunori~B. Hashimoto. 2023.
\newblock Stanford alpaca: An instruction-following llama model.
\newblock \url{https://github.com/tatsu-lab/stanford_alpaca}.

\bibitem[{Touvron et~al.(2023)Touvron, Lavril, Izacard, Martinet, Lachaux, Lacroix, Rozière, Goyal, Hambro, Azhar, Rodriguez, Joulin, Grave, and Lample}]{llama}
Hugo Touvron, Thibaut Lavril, Gautier Izacard, Xavier Martinet, Marie-Anne Lachaux, Timothée Lacroix, Baptiste Rozière, Naman Goyal, Eric Hambro, Faisal Azhar, Aurelien Rodriguez, Armand Joulin, Edouard Grave, and Guillaume Lample. 2023.
\newblock \href {http://arxiv.org/abs/2302.13971} {Llama: Open and efficient foundation language models}.

\bibitem[{Wan et~al.(2023)Wan, Pu, Sun, Garimella, Chang, and Peng}]{wan2023kelly}
Yixin Wan, George Pu, Jiao Sun, Aparna Garimella, Kai-Wei Chang, and Nanyun Peng. 2023.
\newblock \href {http://arxiv.org/abs/2310.09219} {"kelly is a warm person, joseph is a role model": Gender biases in llm-generated reference letters}.

\bibitem[{Wang et~al.(2023{\natexlab{a}})Wang, Liang, Meng, Sun, Shi, Li, Xu, Qu, and Zhou}]{wang2023chatgpt}
Jiaan Wang, Yunlong Liang, Fandong Meng, Zengkui Sun, Haoxiang Shi, Zhixu Li, Jinan Xu, Jianfeng Qu, and Jie Zhou. 2023{\natexlab{a}}.
\newblock \href {http://arxiv.org/abs/2303.04048} {Is chatgpt a good nlg evaluator? a preliminary study}.

\bibitem[{Wang et~al.(2023{\natexlab{b}})Wang, Li, Chen, Zhu, Lin, Cao, Liu, Liu, and Sui}]{wang2023large}
Peiyi Wang, Lei Li, Liang Chen, Dawei Zhu, Binghuai Lin, Yunbo Cao, Qi~Liu, Tianyu Liu, and Zhifang Sui. 2023{\natexlab{b}}.
\newblock \href {http://arxiv.org/abs/2305.17926} {Large language models are not fair evaluators}.

\bibitem[{Wang et~al.(2023{\natexlab{c}})Wang, Kordi, Mishra, Liu, Smith, Khashabi, and Hajishirzi}]{wang2023selfinstruct}
Yizhong Wang, Yeganeh Kordi, Swaroop Mishra, Alisa Liu, Noah~A. Smith, Daniel Khashabi, and Hannaneh Hajishirzi. 2023{\natexlab{c}}.
\newblock \href {http://arxiv.org/abs/2212.10560} {Self-instruct: Aligning language models with self-generated instructions}.

\bibitem[{Wang et~al.(2022)Wang, Mishra, Alipoormolabashi, Kordi, Mirzaei, Naik, Ashok, Dhanasekaran, Arunkumar, Stap, Pathak, Karamanolakis, Lai, Purohit, Mondal, Anderson, Kuznia, Doshi, Pal, Patel, Moradshahi, Parmar, Purohit, Varshney, Kaza, Verma, Puri, Karia, Doshi, Sampat, Mishra, Reddy~A, Patro, Dixit, and Shen}]{wang-etal-2022-super}
Yizhong Wang, Swaroop Mishra, Pegah Alipoormolabashi, Yeganeh Kordi, Amirreza Mirzaei, Atharva Naik, Arjun Ashok, Arut~Selvan Dhanasekaran, Anjana Arunkumar, David Stap, Eshaan Pathak, Giannis Karamanolakis, Haizhi Lai, Ishan Purohit, Ishani Mondal, Jacob Anderson, Kirby Kuznia, Krima Doshi, Kuntal~Kumar Pal, Maitreya Patel, Mehrad Moradshahi, Mihir Parmar, Mirali Purohit, Neeraj Varshney, Phani~Rohitha Kaza, Pulkit Verma, Ravsehaj~Singh Puri, Rushang Karia, Savan Doshi, Shailaja~Keyur Sampat, Siddhartha Mishra, Sujan Reddy~A, Sumanta Patro, Tanay Dixit, and Xudong Shen. 2022.
\newblock \href {https://doi.org/10.18653/v1/2022.emnlp-main.340} {Super-{N}atural{I}nstructions: Generalization via declarative instructions on 1600+ {NLP} tasks}.
\newblock In \emph{Proceedings of the 2022 Conference on Empirical Methods in Natural Language Processing}, pages 5085--5109, Abu Dhabi, United Arab Emirates. Association for Computational Linguistics.

\bibitem[{Wei et~al.(2022)Wei, Bosma, Zhao, Guu, Yu, Lester, Du, Dai, and Le}]{wei2022flan}
Jason Wei, Maarten Bosma, Vincent Zhao, Kelvin Guu, Adams~Wei Yu, Brian Lester, Nan Du, Andrew~M. Dai, and Quoc~V Le. 2022.
\newblock \href {https://openreview.net/forum?id=gEZrGCozdqR} {Finetuned language models are zero-shot learners}.
\newblock In \emph{International Conference on Learning Representations}.

\bibitem[{Yuan et~al.(2021)Yuan, Neubig, and Liu}]{BARTScore}
Weizhe Yuan, Graham Neubig, and Pengfei Liu. 2021.
\newblock \href {https://proceedings.neurips.cc/paper_files/paper/2021/file/e4d2b6e6fdeca3e60e0f1a62fee3d9dd-Paper.pdf} {Bartscore: Evaluating generated text as text generation}.
\newblock In \emph{Advances in Neural Information Processing Systems}, volume~34, pages 27263--27277. Curran Associates, Inc.

\bibitem[{Zhang et~al.(2020)Zhang, Kishore, Wu, Weinberger, and Artzi}]{BERTScore}
Tianyi Zhang, Varsha Kishore, Felix Wu, Kilian~Q. Weinberger, and Yoav Artzi. 2020.
\newblock \href {https://openreview.net/forum?id=SkeHuCVFDr} {Bertscore: Evaluating text generation with bert}.
\newblock In \emph{International Conference on Learning Representations}.

\bibitem[{Zhao et~al.(2023)Zhao, Strube, and Eger}]{zhao-etal-2023-discoscore}
Wei Zhao, Michael Strube, and Steffen Eger. 2023.
\newblock \href {https://aclanthology.org/2023.eacl-main.278} {{D}isco{S}core: Evaluating text generation with {BERT} and discourse coherence}.
\newblock In \emph{Proceedings of the 17th Conference of the European Chapter of the Association for Computational Linguistics}, pages 3865--3883, Dubrovnik, Croatia. Association for Computational Linguistics.

\bibitem[{Zhou et~al.(2023)Zhou, Liu, Xu, Iyer, Sun, Mao, Ma, Efrat, Yu, Yu, Zhang, Ghosh, Lewis, Zettlemoyer, and Levy}]{zhou2023lima}
Chunting Zhou, Pengfei Liu, Puxin Xu, Srini Iyer, Jiao Sun, Yuning Mao, Xuezhe Ma, Avia Efrat, Ping Yu, Lili Yu, Susan Zhang, Gargi Ghosh, Mike Lewis, Luke Zettlemoyer, and Omer Levy. 2023.
\newblock \href {http://arxiv.org/abs/2305.11206} {Lima: Less is more for alignment}.

\end{thebibliography}
